%% file: main.tex
\documentclass[10pt,leqno]{amsart}

\usepackage{cite}
\usepackage[dvipdfmx]{graphicx}
\usepackage{caption}
\usepackage{subfig}
\usepackage{comment}
\usepackage{indentfirst,csquotes}
\usepackage{amssymb,amsthm,amsmath}
\usepackage{xcolor,paralist,hyperref,titlesec,fancyhdr,etoolbox}
\usepackage{lipsum}

\titleformat{\section}
  {\normalfont\Large\bfseries}
  {\thesection.}{1em}{}
\titlespacing*{\section}{0pt}{2ex plus 1ex minus .2ex}{1ex plus .2ex}

\titleformat{\subsection}
  {\normalfont\large\bfseries}
  {\thesubsection}{1em}{}
\titlespacing*{\subsection}{0pt}{1.5ex plus 1ex minus .2ex}{1ex plus .2ex}

\hypersetup{
  colorlinks=true,
  linkcolor=black,
  filecolor=black,
  urlcolor=black
}

\begin{document}

\title{Evaluation of an Autonomous Surface Robot Equipped with a Transformable Mobility Mechanism for Efficient Mobility Control}

\author[Y. Fujii et al.]{Yasuyuki Fujii, Dinh Tuan Tran and Joo-Ho Lee}
\address{College of Information Science and Engineering, Ritsumeikan University, Japan}
\email{fujiyasu@fc.ritsumei.ac.jp}
\date{\today}

\maketitle


\begin{abstract}
Efficient mobility and power consumption are critical for autonomous water surface robots in long-term water environmental monitoring.
This study develops and evaluates a transformable mobility mechanism for a water surface robot with two control modes: station-keeping and traveling to improve energy efficiency and maneuverability.
Field experiments show that, in a round-trip task between two points, the traveling mode reduces power consumption by 10\% and decreases the total time required for travel by 5\% compared to the station-keeping mode.
These results confirm the effectiveness of the transformable mobility mechanism for enhancing operational efficiency in patrolling on water surface.
\end{abstract}

\bigskip

\section{INTRODUCTION}
Recent environmental changes have significantly affected aquatic ecosystems, increasing the demand for continuous monitoring.
We have been developing autonomous water surface robots capable of moving to arbitrary positions and maintaining station-keeping for water environment monitoring\cite{fujii2024multi,fujii2022evaluation,fujii2022efficient,fujii2020development}.
The robot conducts a patrolling task, where it travels between multiple observation points and maintains its position at each point for data collection before moving to the next location.
In previous research, we proposed an 'X'-shaped robot, and the results showed that power consumption was significantly higher in the task of traveling between points than in station-keeping\cite{fujii2024multi}.
To improve efficiency, this study develops a transformable mobility mechanism that enables efficient station-keeping and traveling.
Its effectiveness is validated through field experiments by comparing power consumption and total travel time over a fixed distance under different modes.

\section{BIWAKO-8}
\subsection{Structure}
In this study, we developed a small autonomous water surface robot named BIWAKO-8 (Brief Intelligent WAter Knowledge Observer).  
Figure~\ref{fig:biwako_modes} illustrates the developed BIWAKO-8.
Figures~\ref{fig:keep} and \ref{fig:straight} illustrate BIWAKO-8 in station-keeping mode and traveling mode, respectively.
BIWAKO-8 is composed of two main components: the computer module and the thruster mount module.
The computer module contains a battery, flight controller, control computer, and servo controller, all of which are housed inside a waterproof chamber.  
The flight controller is used for autonomous navigation and is equipped with a 9 axis inertial sensor and a GNSS receiver, providing essential modules for the autonomous movement of BIWAKO-8 on the water surface.  
Figure~\ref{fig:thruster_arm} illustrates a thruster arm of the thruster mount module.
Each thruster arm consists of a servo motor for joint angle control, another servo motor for thruster angle adjustment, and a thruster for propulsion.
The dimensions of BIWAKO-8 are as follows. When fully extended, the maximum length of the link arms is 1.2 m, while the height, excluding the antenna, is 0.35 m. The total weight is 5.5 kg, making it portable for a single person.

\input{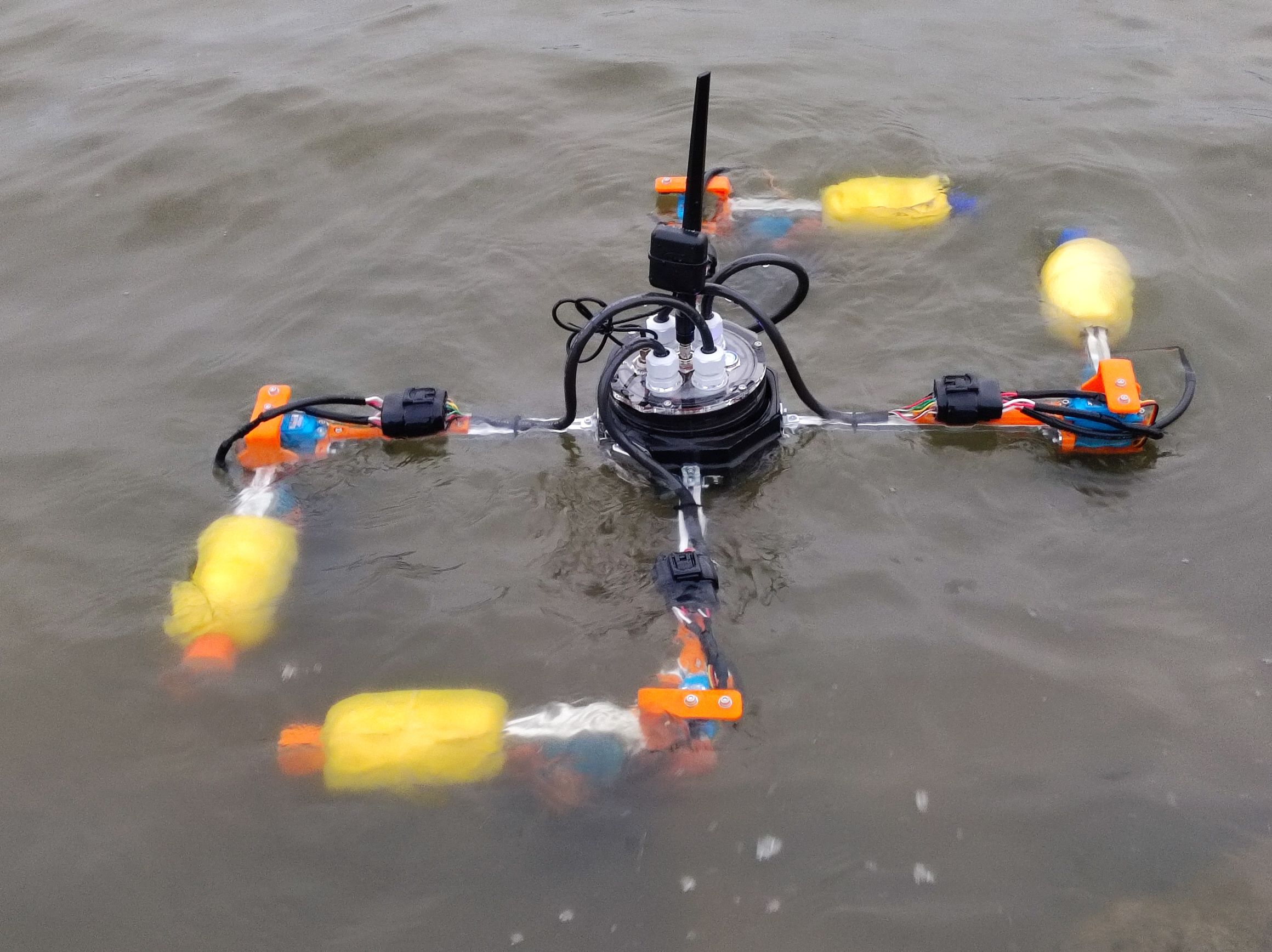}
\input{thruster_arm}

\subsection{Transformable Mobility Mechanism}
\input{mode}
The transformable mobility mechanism is designed to adjust the thruster configuration and shape of BIWAKO-8.  
Optimal thruster mounting angles and shape configuration play a critical role in ensuring efficient operation in missions alternating between station-keeping and traveling.
According to~\cite{tanakitkorn2022impacts}, optimal thruster configurations have been analyzed for different tasks.

Figure~\ref{fig:mode} illustrates the thruster angles, link angles, and movement direction of the robot in both station-keeping mode and traveling mode.
The key feature of station-keeping mode is its maneuverability to move instantly in all directions.
This feature allows the robot to efficiently maintain its position in environments where disturbances are applied from various directions.
However, a disadvantage of this mode is the significant energy loss incurred during traveling.
On the other hand, the key feature of the traveling mode is its ability to travel efficiently between target positions.
In this mode, the thrusters are directly oriented in the direction of movement, minimizing energy loss.
Additionally, the robot transforms into a shape that reduces resistance in the direction of travel, further improving energy efficiency.
However, traveling mode is less suitable for maintaining its position in environments with disturbances from arbitrary directions.
By switching between these two modes, the robot can efficiently patrol between decided target observation points in real field.

\section{Field Experiment}
\subsection{Experimental Conditions}
To verify the effectiveness of the transformable mobility mechanism, BIWAKO-8 was operated in a real environment under two different control modes.
The robot was given two predefined target positions and moved back and forth between them.  
In this experiment, the distance between the two points was set at 20 m, and the robot was programmed to travel a total distance of 200 m by completing ten round trips.
The evaluation criteria for the field experiment were the average lap time per round trip, the total power consumption, and the total travel time.

\subsection{Experimental Results}
Figure \ref{fig:laptime} shows the lap time per round trip for each control mode.  
The average lap time was 104.0 seconds in the station-keeping mode and 99.2 seconds in the traveling mode.
Figure \ref{fig:power_consumption} presents the time series graph of power consumption on ten round trips in both modes.  
In terms of cumulative power consumption, the total energy used was 516.1 kJ in the station-keeping mode and 466.7 kJ in the traveling mode, achieving a 10\% reduction in total power consumption.  
The total time required for ten round trips was 1039.7 seconds in the station-keeping mode and 991.9 seconds in the traveling mode, demonstrating a 5\% reduction in travel time.
These results confirm that the traveling mode is more efficient than the station-keeping mode for moving between multiple observation points.
By switching between the station-keeping mode for maintaining position and the traveling mode for moving between points, the robot can efficiently patrol and complete its missions.

\input{lap}
\input{power_consumption}

\section{Conclusion}
This paper examined the effectiveness of a transformable mobility mechanism for a small autonomous water surface robot designed for efficient patrol missions that involve alternating between station-keeping and traveling.
The mechanism was evaluated through real-world field experiments, demonstrating improvements in both energy consumption and travel time.
Future work will focus on conducting additional power consumption measurements under mission scenarios that involve alternating between station-keeping and traveling to further validate the robot's efficiency in patrolling observations.

$\,$

$\,$

\end{document}

%% file: BIWAKO-8.tex
\begin{figure}[tb]
  \centering
  \subfloat[Station-keeping mode]{%
    \includegraphics[width=0.45\linewidth]{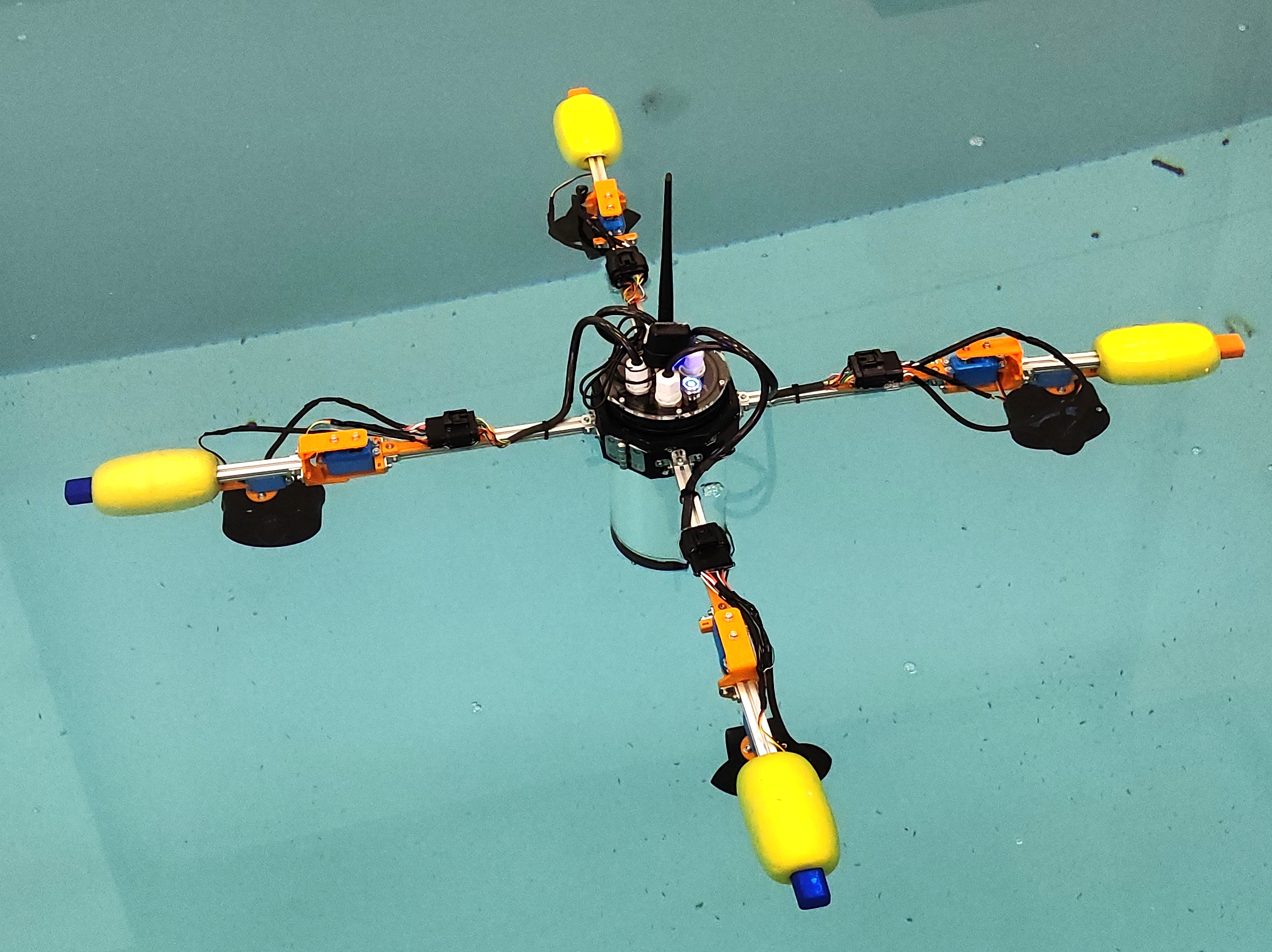}
    \label{fig:keep}
  }
  \hspace{0.02\linewidth} 
  \subfloat[Traveling mode]{%
    \includegraphics[width=0.45\linewidth]{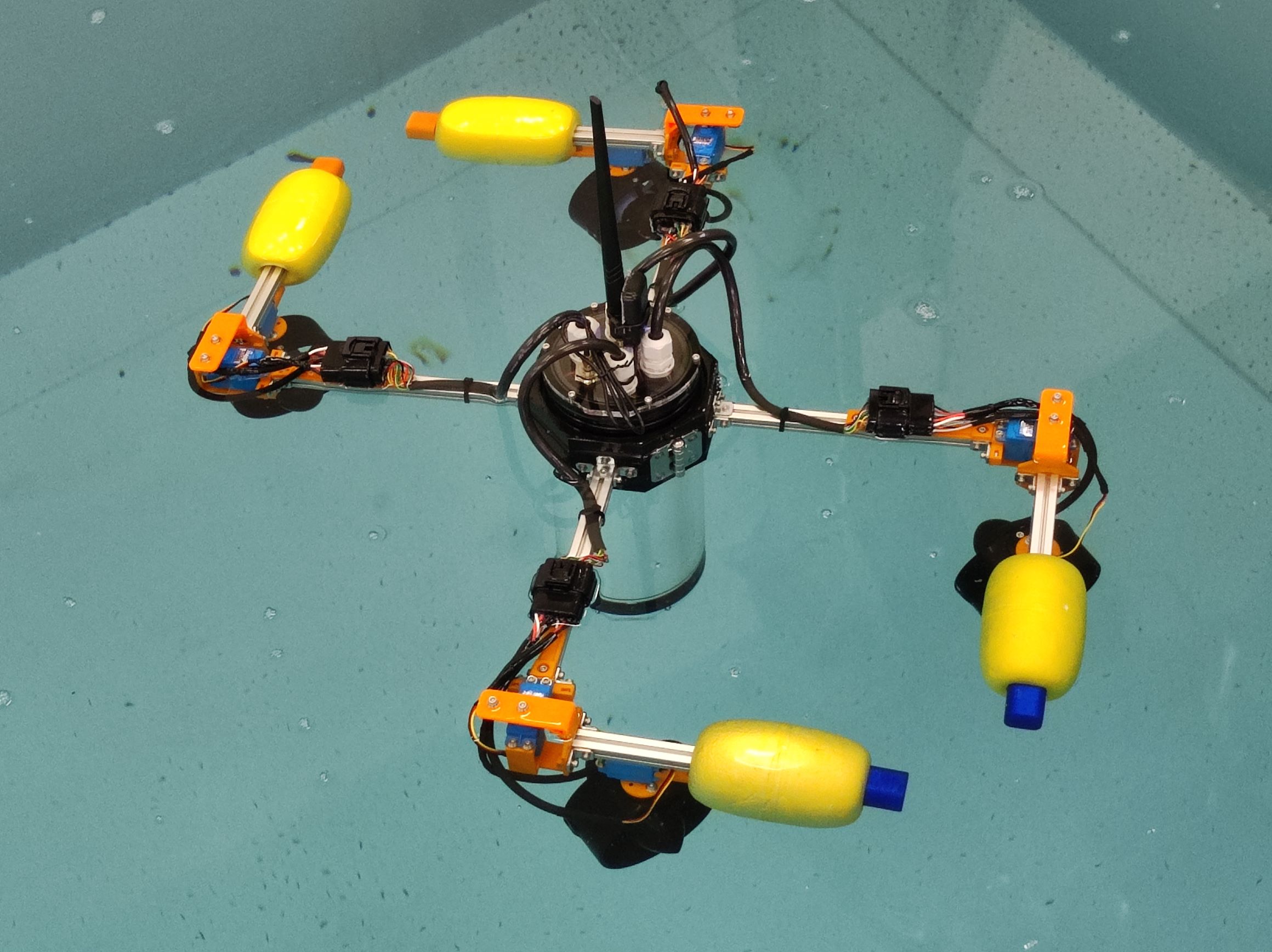}
    \label{fig:straight}
  }
  \caption{BIWAKO-8}
  \label{fig:biwako_modes}
\end{figure}

%% file: thruster_arm.tex
\begin{figure}[tb]
  \centering
    \includegraphics[width=0.6\linewidth]{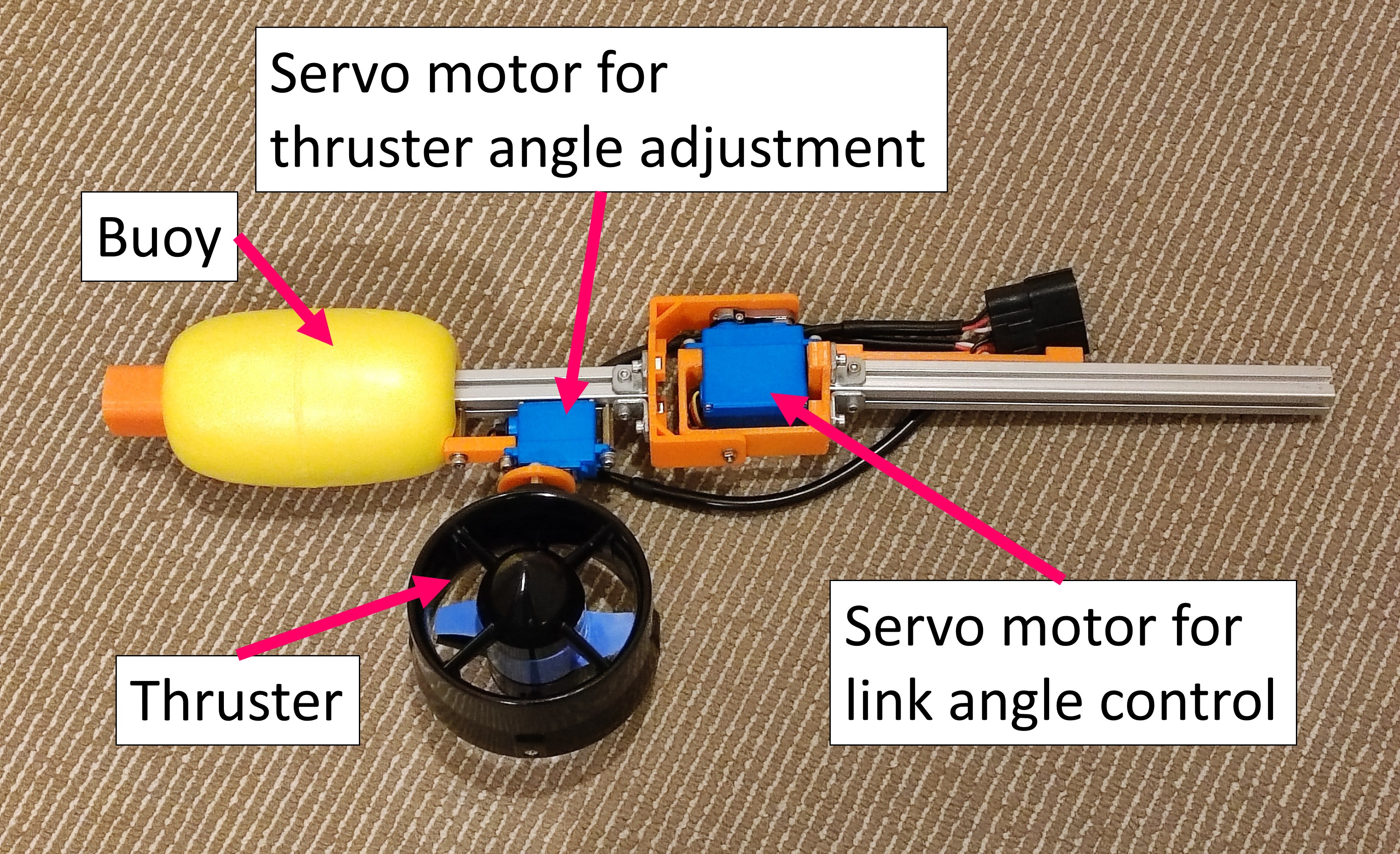}
    \caption{Thruster arm}
    \label{fig:thruster_arm}
\end{figure}

%% file: mode.tex
\begin{figure}[tb]
  \centering
    \includegraphics[width=0.7\linewidth]{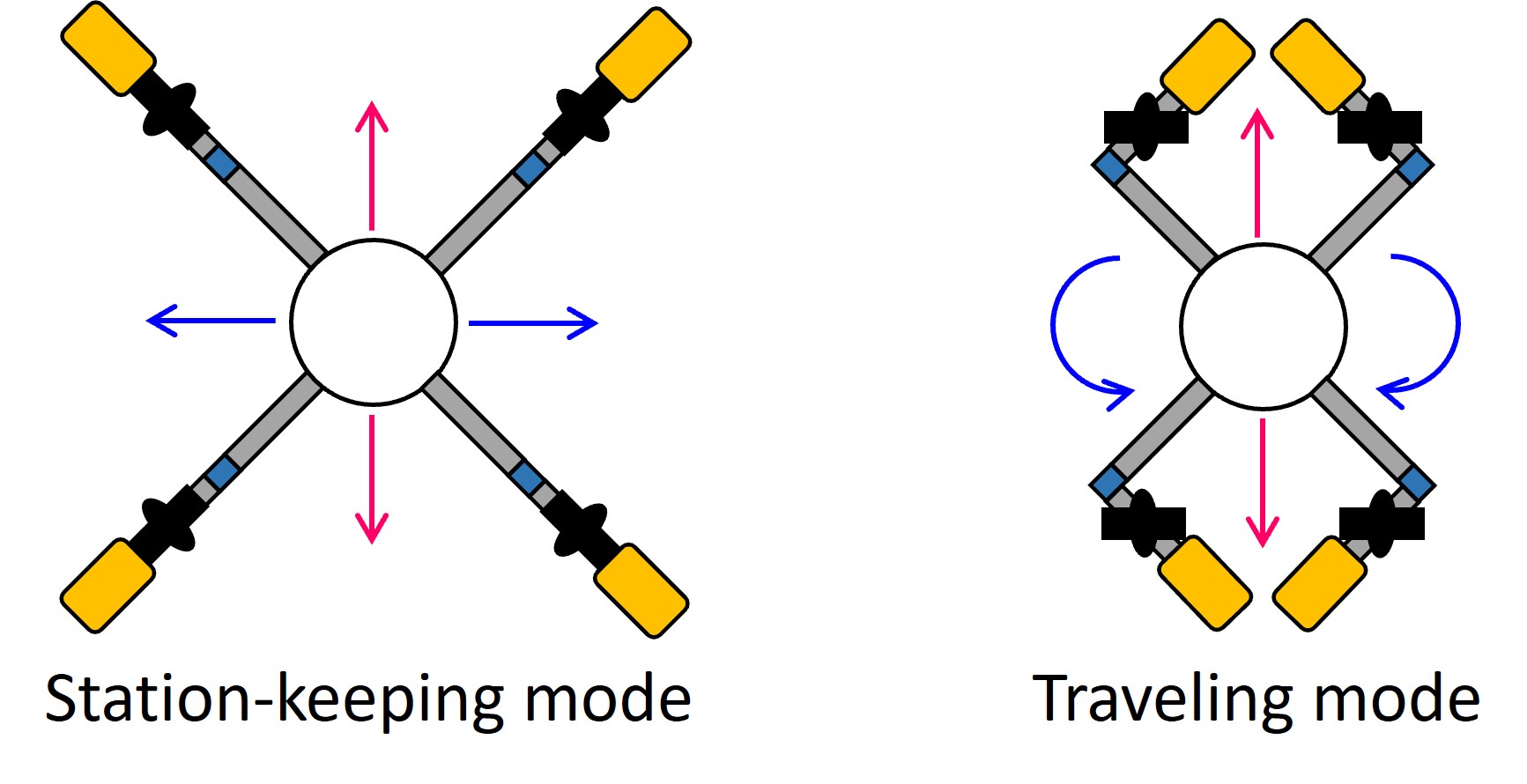}
    \caption{Configuration of the mobility mechanism in station-keeping and patrolling modes}
    \label{fig:mode}
\end{figure}

%% file: lap.tex
\begin{figure}[t]
  \centering
    \includegraphics[width=0.8\linewidth]{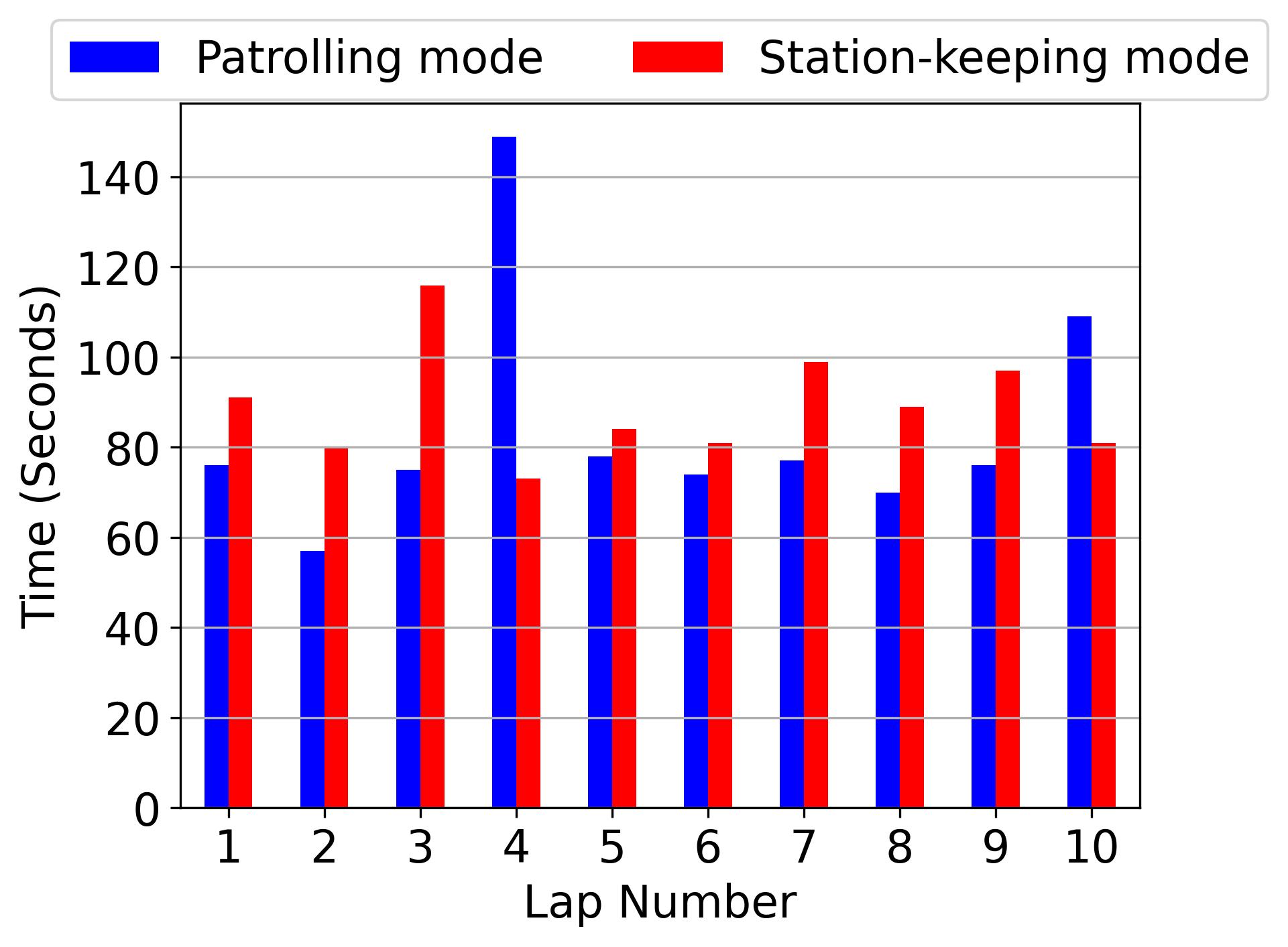}
    \caption{Lap time}
    \label{fig:laptime}
\end{figure}

%% file: power_consumption.tex
\begin{figure}[t]
  \centering
    \includegraphics[width=\linewidth]{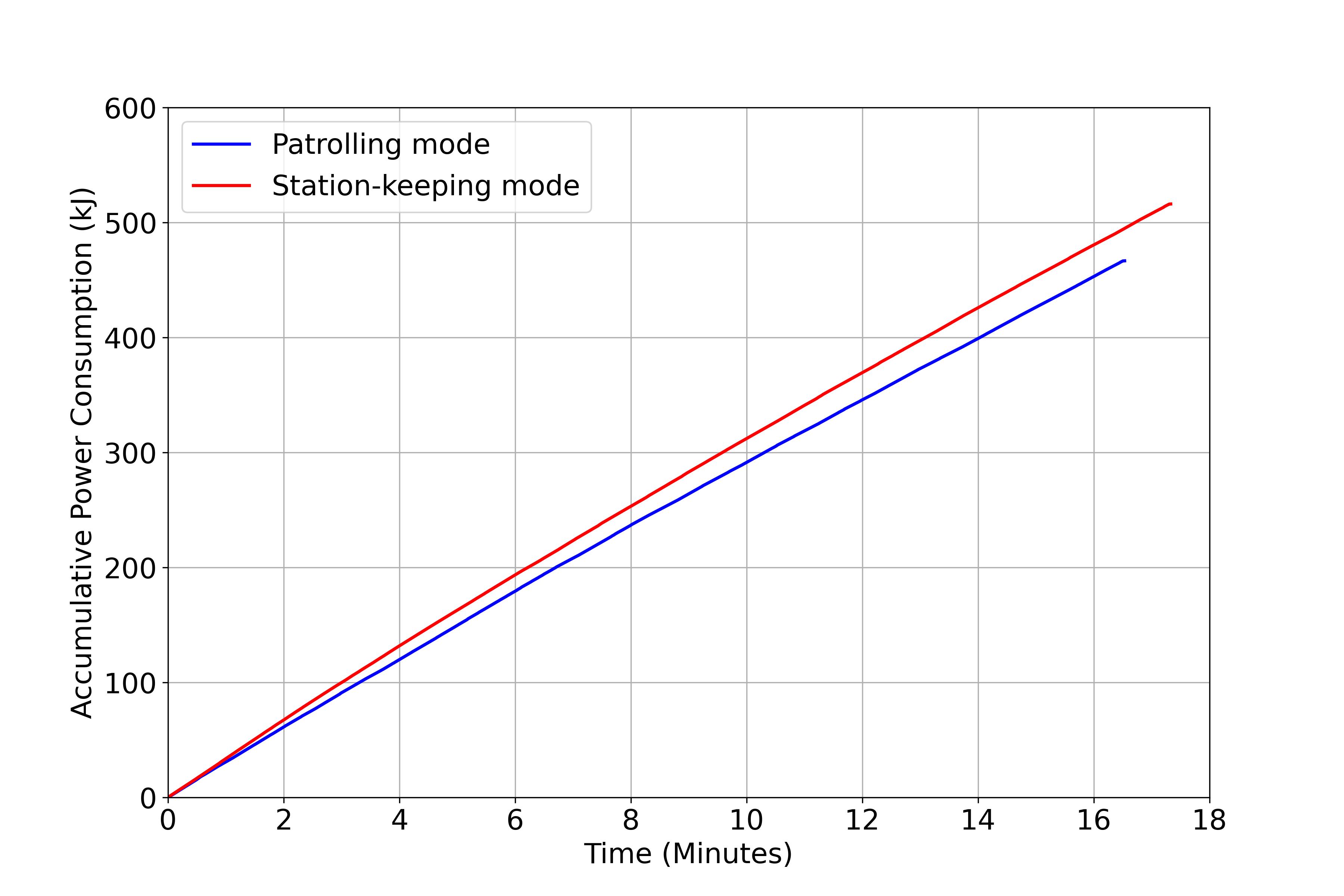}
    \caption{Power consumption result}
    \label{fig:power_consumption}
\end{figure}